%% file: main.tex
\def\year{2022}\relax
\documentclass[letterpaper]{article} 
\usepackage{aaai22}  
\usepackage{times}  
\usepackage{helvet}  
\usepackage{courier}  
\usepackage[hyphens]{url}  
\usepackage{graphicx} 
\usepackage{natbib}  
\usepackage{caption} 
\DeclareCaptionStyle{ruled}{labelfont=normalfont,labelsep=colon,strut=off} 
\usepackage{algorithm}
\usepackage{algorithmic}
\usepackage[namelimits]{amsmath}
\usepackage{amssymb}
\usepackage{bm}

\usepackage[utf8]{inputenc} 
\usepackage{url}            
\usepackage{booktabs}       
\usepackage{amsfonts}       
\usepackage{nicefrac}       
\usepackage{microtype}      
\usepackage{xcolor}         
\usepackage{soul}

\usepackage{newfloat}
\usepackage{listings}

\usepackage{subfigure}

\floatstyle{ruled}
\newfloat{listing}{tb}{lst}{}
\floatname{listing}{Listing}
\title{Task-Customized Self-Supervised Pre-training with Scalable Dynamic Routing}

\author {
    Zhili Liu\textsuperscript{\rm 1,2},
    Jianhua Han\textsuperscript{\rm 2},
    Lanqing Hong\textsuperscript{\rm 2},
    Hang Xu\textsuperscript{\rm 2},
    Kai Chen\textsuperscript{\rm 1},
    Chunjing Xu\textsuperscript{\rm 2},
    Zhenguo Li\textsuperscript{\rm 2}
}
\affiliations {
    \textsuperscript{\rm 1}  Department of Computer Science and Engineering, \\Hong Kong University of Science and Technology\\
    \textsuperscript{\rm 2} Huawei Noah’s Ark Lab\\
    \{zhili.liu,  kai.chen\}@connect.ust.hk,
    \{hanjianhua4, honglanqing, xu.hang, xuchunjing, li.zhenguo\}@huawei.com

}

\begin{document}

\maketitle

\begin{abstract}
Self-supervised learning (SSL), especially contrastive methods, has raised attraction recently as it learns effective transferable representations without semantic annotations.
A common practice for self-supervised pre-training is to use as much data as possible.
For a specific downstream task, however, 
involving irrelevant data in pre-training may degenerate the downstream performance, observed from our extensive experiments.
On the other hand, for existing SSL methods, it is burdensome and infeasible to use different downstream-task-customized datasets in pre-training for different tasks.
To address this issue, we propose a novel SSL paradigm called Scalable Dynamic Routing (SDR), which can be trained once and deployed efficiently to different downstream tasks with task-customized pre-trained models. 
Specifically, we construct the SDRnet with various sub-nets and train each sub-net with only one subset of the data by data-aware progressive training.
When a downstream task arrives, we route among all the pre-trained sub-nets to get the best along with its corresponding weights.
Experiment results show that our SDR can train 256 sub-nets on ImageNet simultaneously, which provides better transfer performance than a unified model trained on the full ImageNet,
achieving state-of-the-art (SOTA) averaged accuracy over 11 downstream classification tasks and AP on PASCAL VOC detection task. 
\end{abstract}

\input{intro}

\input{related}
\input{background}

\input{method}
\input{experiment}
\input{conclusion}
\newpage
{
\small \bibliography{Bib}
}

\clearpage
\appendix
\section*{Supplementary Materials: Task-customized Self-supervised Pre-training with Scalable Dynamic Routing}
\input{appendixA}
\input{appendixB}
\input{appendixC}
\input{appendixD}

\end{document}

%% file: intro.tex
\section{Introduction}

Self-supervised learning (SSL) has attracted lots of attention recently~\citep{caron2020unsupervised, he2020momentum, byol20},
which learns representations via pretext tasks without semantic annotations.
Recent works in SSL~\citep{xu2020hierarchical,chen2021multisiam} show competitive or even better performance on various downstream tasks compared with supervised learning.
Without the need of annotation, SSL makes it possible to use a large amount of unlabeled data (e.g., YFCC100M~\citep{tian2021divide}, Instagram~\citep{goyal2021self}
and SODA10M~\citep{han2021soda10m})
in model pre-training.
However, will more data in self-supervised pre-training always lead to better transfer performance?
In other words, \textit{for a specific downstream task, will irrelevant data in pre-training hurt the downstream performance instead?}

\begin{figure*}[t!]
    \centering
    \subfigure[Samples from Subset-A/B]{\includegraphics[height=3.2cm]{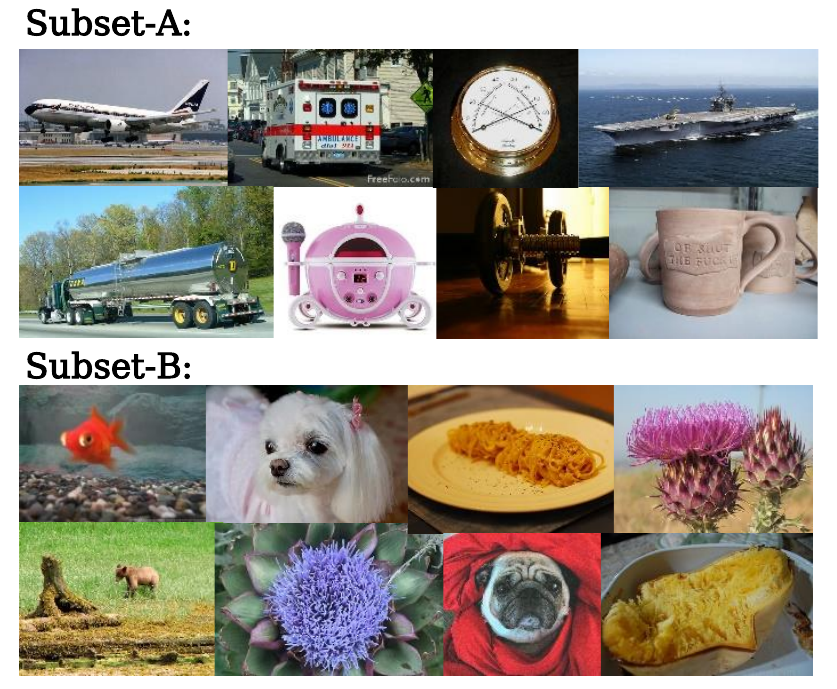}\label{fig:sample}}
    \hspace{3mm}
    \subfigure[Performance of models pre-trained on different datasets]{\includegraphics[height=3.2cm]{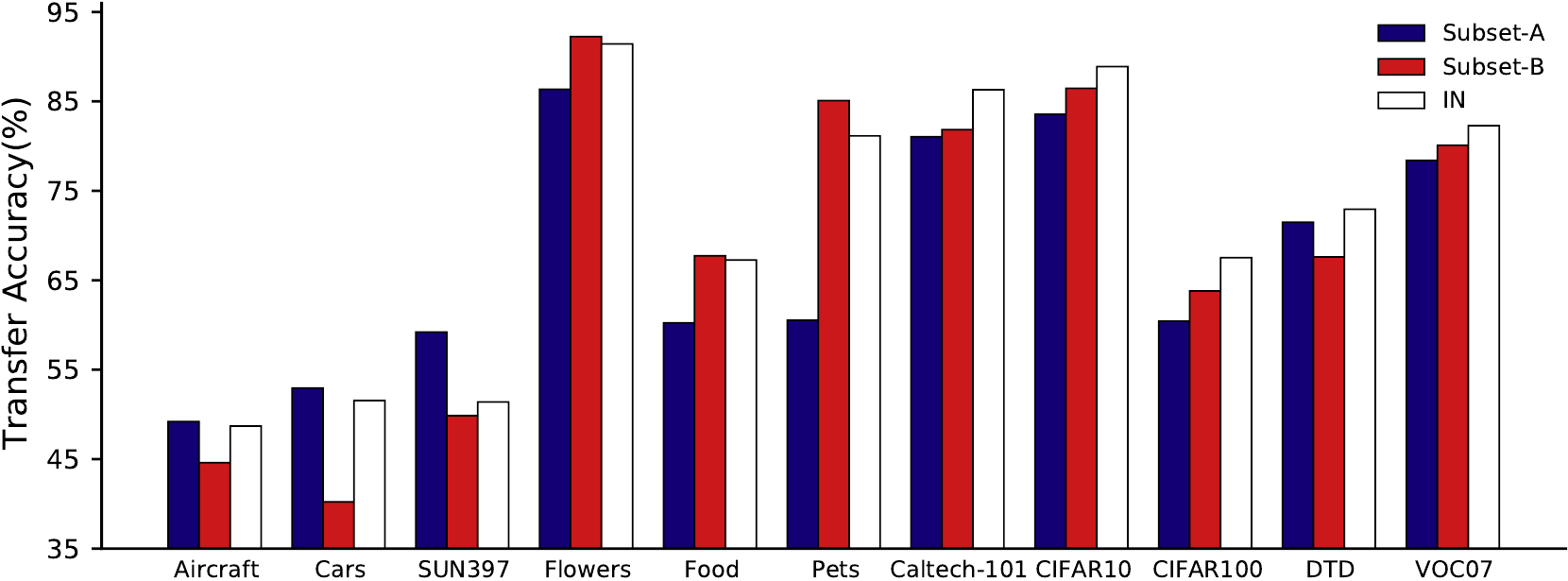}\label{fig:pre acc}}
    \caption{Transfer performance for models pre-trained on ImageNet Subset-A, ImagNet Subset-B and the full ImageNet on different downstream datasets. 
    (a) Subset-A is occupied by inanimate objects mostly, while Subset-B mainly contains organisms;
    (b) The model pre-trained on the full ImageNet only has the best performance on five out of the eleven tasks.}
    \label{fig:splits}
\end{figure*}

To answer the above questions, we first conduct a preliminary experiment in Sec.~\ref{sec:pre} to evaluate the transfer performance of SSL models pre-trained on datasets with different semantics.
We deliberately split the ImageNet into two disjoint subsets, namely Subset-A and Subset-B, based on their semantic dissimilarity in WordNet Tree~\citep{miller1998wordnet}.
We pre-train models with Subset-A, Subset-B and the full ImageNet separately using SimSiam~\citep{chen2020exploring} without data annotations and evaluate the transfer performance on 11 downstream classification datasets.
The training epochs for the three models are the same.
As shown in Fig.~\ref{fig:pre acc}, the model pre-trained on Subset-A shows the best transfer performance on Aircraft, Cars and SUN397, while the model pre-trained on Subset-B performs the best on Flowers, Pets, and Food.
Only five out of eleven downstream tasks benefit more from the full ImageNet. 
The results indicate that involving irrelevant data in pre-training might instead hurt the downstream performance.
This phenomenon is identified as the \emph{negative transfer} in self-supervised pre-training.
Similar observations have also been discussed in \citep{cole2021does} and \citep{tian2021divide}.
\citep{cole2021does} further investigate the importance of using semantic-similar data in model pre-training for better transfer performance. 

Prevailing SSL methods, such as MoCo-v2~\citep{chen2020improved} and SimSiam~\citep{chen2020exploring}, usually neglect the influence of negative transfer and provide a common pre-trained model for different downstream tasks.
A naive extension to eliminate the effects of negative transfer is to pre-train models with task-customized datasets.
However, such an extension is actually impractical considering the burdensome computational cost of pre-training.
\cite{tian2021divide} simply splits a large-scale dataset (i.e., YFCC100M) into different subsets for customized model pre-training, which is not scalable for a large number of downstream tasks.
It is desirable to develop an efficient SSL paradigm that provides task-customized pre-training models.

In this work, we propose a novel SSL paradigm called Scalable Dynamic Routing (SDR), which achieves dynamic pre-training and efficient deployment for different downstream tasks.
Specifically, we construct the SDRnet with various sub-nets and train each sub-net with different subsets of the data, which contain different semantic clusters.
We further propose a data-aware progressive training framework to stabilize the pre-training procedure of sub-nets and avoid collapse.
When a downstream task arrives, we route among all sub-nets to obtain the best pre-trained model along with its weights.
By using SDR, we are able to pre-train a series of sub-nets simultaneously for the efficient deployment of various downstream tasks.
To summarize, our main contributions are:
\begin{itemize}
    \item With extensive experiments, we identify the negative transfer phenomenon in SSL that pre-training with irrelevant data might degenerate the transfer performance in specific downstream tasks.

    \item We propose Scalable Dynamic Routing (SDR), a novel SSL paradigm that can alleviate the effects of negative transfer by providing efficient and scalable task-customized self-supervised pre-training models.

    \item We successfully train 256 sub-nets simultaneously on ImageNet and achieve the state-of-the-art averaged accuracy among 11 downstream classification datasets and AP on PASCAL VOC detection task.
\end{itemize}

%% file: related.tex
\section{Related work}

\textbf{Self-supervised learning,}
especially contrastive learning, learns representations without data annotation by ``learning to compare'' through a Noise Contrastive Estimation (NCE)~\cite{he2020momentum} objective.
Recently, instance-instance contrastive learning ~\citep{wu2018unsupervised,he2020momentum,chen2020simple}  becomes prevailing, which directly studies the relationships between representations of different samples.
BYOL~\citep{byol20} and SimSiam~\citep{chen2020exploring} further claims meaningful representations can be learned without (i) negative sample pairs, (ii) large batches, and (iii) momentum encoders.
Besides, clustering-based methods, including PCL-v1, PCL-v2~\citep{li2020prototypical},
and SwAV~\cite{caron2020unsupervised}, leverage clustering to yield pseudo labels for learning representations.
However, existing SSL methods usually offer a unified pre-trained model which may not be applicable for various downstream tasks when negative transfer occurs, as shown in Sec.~\ref{sec:pre}.
It is impractical to pre-train different SSL models for different tasks due to the burdensome computational cost, thus is desirable to develop an efficient and scalable task-customized SSL paradigm.

\noindent\textbf{Dynamic neural network} 
is an emerging topic~\cite{han2021dynamic}. 
Unlike static networks with fixed computational graphs and weights during inference, dynamic networks can adapt their structures or parameters to different scenarios, leading to advantages in terms of efficiency, adaptiveness, and performance. 
Based on the dynamic nature, they can be categorized into instance-wise \cite{li2017not,figurnov2017spatially}, spatial-wise \cite{cao2019seernet,wang2019elastic} and temporal-wise networks \cite{campos2017skip,hansen2019neural,tao2019skipping}. 
In order to allow the adaptiveness of our pre-trained backbone to different tasks and datasets, we need to explore task-wise/dataset-wise dynamic networks. 
Compared with instance-wise dynamic networks \cite{odena2017changing,liu2018dynamic}, 
our method focuses on selecting the best candidate model for a downstream task/dataset and fixes the network structure during inference.

\noindent\textbf{Multi-task Learning}
(MTL) aims at learning a model that can perform well on several downstream tasks, which are usually pre-defined during training, while SDR can not foresee any downstream tasks when pre-training.
\citep{mcdermott2021comprehensive}, \citep{liu2019multi} and \citep{hu2019strategies} show that the model using a shared backbone for all tasks and multi-heads for different specific tasks, namely hard-parameter sharing, is useful on time-series data, language and graph data separately.
\citep{gao2021network} shows that network design can better benefit the task relationship, while \citep{gao2021network} trains a mask along with the model parameters, so each task has its own mask.
SDR is designed differently by super-sub-net structure, neither requiring multi-heads nor masks, making SDR more parameter-efficient. 
Furthermore, SDR is also scalable for training 256 sub-tasks simultaneously, which is significantly larger than most MTL methods.

%% file: background.tex
\section{Preliminary on Negative Transfer}
\label{sec:pre}

In this section, we conduct a preliminary experiment to evaluate the transfer performance of models pre-trained on datasets with different semantic annotations.
Following~\cite{huh2016makes}, we split the ImageNet into two disjoint subsets, namely Subset-A and Subset-B, based on their semantic dissimilarity in WordNet Tree~\citep{miller1998wordnet},
which can be achieved by searching the WordNet hierarchy to avoid two splits having the same ancestor at depth four.

In this case, classes in Subset-A are sufficiently disjoint from Subset-B.
Specifically, images in Subset-A are primarily inanimate objects, such as cars and airplanes, while Subset-B mainly contains organisms, such as plants and animals.
See Fig.~\ref{fig:sample} as an illustration.

Then, we pre-train with Subset-A, Subset-B and the full ImageNet separately using SimSiam~\citep{chen2020exploring} without data annotations, and evaluate the transfer performance on 11 downstream classification datasets via the many-shot classification protocol following~\cite{ericsson2020well}. 
See more experimental details and hyper-parameters in Appendix A. 

The results are summarized in Fig.~\ref{fig:pre acc}. 
As can be seen, the model pre-trained on Subset-A shows the best performance on Aircraft, Cars and SUN397. 
Specifically, for SUN397, the model with Subset-A results in a 7.83\% improvement on classification accuracy compared with the model pre-trained on the full ImageNet.
On the other hand, the model pre-trained on Subset-B performs the best on Flowers, Pets, and Food.
These results are consistent with the observations that Subset-A is mostly inanimate objects, while Subset-B mainly contains organisms.
Only five out of the eleven downstream tasks benefit from the full ImageNet, suggesting that more data in pre-training is not always better.
Involving semantic-irrelevant data in pre-training might hurt the downstream-task performance.
The observation of \emph{negative transfer} in self-supervised pre-training motivates us to develop an efficient but scalable task-customized SSL paradigm.

%% file: method.tex
\section{Method}

\begin{figure*}[t] 
	\centering
	\includegraphics[width=1.8\columnwidth]{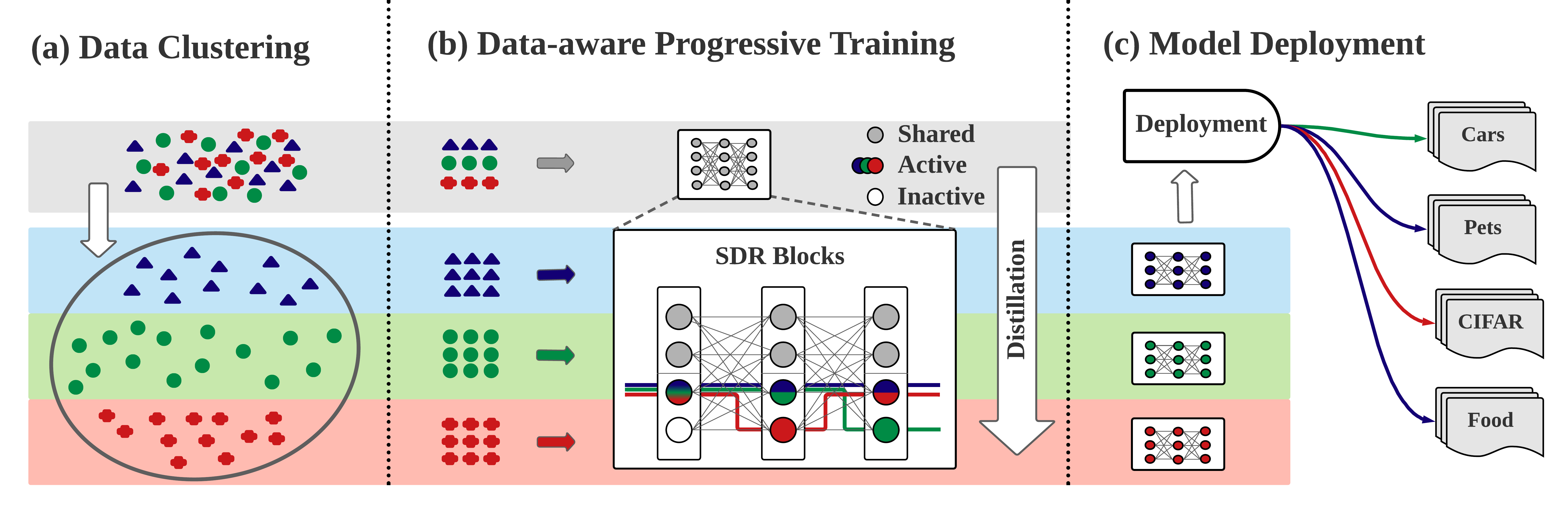}
	\caption{An overview of our proposed SDRnet. 
	(a) We first separate unlabeled images into different subsets by clustering;
	(b) SDRnet is then constructed with various sub-nets and each sub-net is trained with only one subset of the data by data-aware progressive training;
	(c) When a downstream task arrives, we route among all the sub-nets to get the best pre-trained model.}
	\label{fig:blockdes}
\end{figure*}

In this section, we start by a brief introduction of the SimSiam~\citep{chen2020simple}, our simple yet effective SSL baseline, in Sec.~\ref{sec:SSL baseline}.
Then we introduce the proposed Scalable Dynamic Routing (SDR) paradigm for the simultaneous pre-training of a series of sub-nets in Sec.~\ref{sec:SDR}.
Finally, we discuss the efficient deployment of these sub-nets to different downstream tasks in Sec.~\ref{sec:dep}.

\subsection{Overview of SimSiam}
\label{sec:SSL baseline}

SimSiam~\citep{chen2020simple} takes two randomly augmented views $x_1$ and $x_2$ from an image $x$ as inputs.
The two views are processed by an encoder  $f_\theta$, which contains a backbone and a projection MLP head.
The encoder $f_\theta$ shares weights between $x_1$ and $x_2$.
Furthermore, a prediction MLP head $h_\theta$ transforms the output of one view and matches it with the other.
SimSiam learns representations by comparing similarity of the encoder output $f_\theta(\cdot)$ and the prediction head output $h_\theta(\cdot)$.
Finally, a consistency loss is calculated as:

\begin{equation}
        \mathcal{L}_{SSL}(D;\theta) = 
    \mathop{\mathbb{E}}_{\substack{x_1, x_2\sim\tau(x)\\ x \sim D}}
    \frac{h_\theta(f_\theta(x_1))}
    {\|h_\theta(f_\theta(x_1))\|_2} 
    \frac{f_\theta(x_2)}{\|f_\theta(x_2)\|_2},
\end{equation}
where $\|\cdot\|_2$ denotes the $l_2$-norm, and $D$, $\tau(\cdot)$ indicate the unlabeled training dataset and distribution of data augmentation respectively. 
Moreover, the stop-gradient operation is adopted to avoid collapse solutions in the implementation.

\subsection{Scalable Dynamic Routing}
\label{sec:SDR}
As shown in Fig.~\ref{fig:blockdes}, our Scalable Dynamic Routing (SDR) paradigm consists of three steps. 
First, we cluster the dataset into disjoint subsets, then we construct the SDRnet model containing many sub-nets and dynamically train each sub-net with its corresponding subsets through data-aware progressive training. 
Refer to Algorithm 1 in Appendix D for the entire training procedure. 
After pre-training, we can route among all sub-nets to find the one that transfers best to a specific downstream task.
Following are the details.

\textbf{Data clustering.} 
The basic idea of SDR is to apply data with different semantics to train different networks simultaneously and efficiently. 
A clustering procedure is adopted to group the unlabeled training data into different semantic clusters. 
We first pre-train a SimSiam model using the entire dataset and collect all images features, denoted as $\bm{F} = [\bm{f_1},\bm{f_2},...,\bm{f_n}]$. 
Large-scale clustering is performed on \textbf{fixed} $\bm{F}$ following~\cite{caron2020unsupervised}. 
Specifically, we set $k$ to be our desired number of clusters and define the learnable centroids of the clusters as $\bm{C}=[\bm{c_1}, \bm{c_2}, ..., \bm{c_k}]$. Then the assignment of features to the clusters can be computed as $\bm{S} = \bm{F}^T \bm{C}$. We define an auxiliary matrix $\bm{U}=[\bm{u_1},\bm{u_2},...,\bm{u_n}]$, which can be regarded as the posterior distribution of clustering~\cite{asano2019self}. Our goal is to maximize the similarity between $\bm{U}$ and $\bm{S}$, which can be denoted as follows,
\begin{equation}
    \max_{\bm{U}}  [Tr(\bm{U}^T\bm{S}) + \epsilon H(\bm{U})]
    \label{eqn:clustering},
\end{equation}
where $H(\bm{U})$ denotes the entropy of $\bm{U}$. 
We optimize $\bm{U}$ and $\bm{C}$ iteratively. 
$\bm{U}$ is solved by the iterative Sinkhorn-Knopp algorithm~\cite{cuturi2013sinkhorn}, 
while $\bm{C}$ is learned through SGD to minimize the cross entropy between $\bm{U}$ and $\bm{S}=\bm{F}^T \bm{C}$. 
After several epochs of training, we adopt $\bm{S}$ to be our assignment matrix. 
The final clustering result is denoted as $D_i(i=1,2, ..., k)$, and $D_0$ represents the entire dataset.

\textbf{Framework optimization.}
The whole SDRnet will be trained by the entire training set $D_0$, while the $i$-th sub-net will be additionally trained with its corresponding sub-dataset $D_i$. 
Let $W_0$ be the weights of the total network, and $W_i \subseteq W_0 (i = 1, \cdots, k)$ is the weights corresponding to the $i$-th sub-net. 
The training loss can be formalized as:
\begin{equation}
\min_{W_0} [\mathcal{L}_{SSL}(D_0;W_0) + \sum_{i} \mathcal{L}_{SSL}(D_i;W_i)],
\end{equation}
and the overall objective optimizes the weights of the SDRnet and sub-nets simultaneously on their corresponding datasets.

\begin{figure}[h] 
	\centering
	{\subfigure[Data-aware progressive training at phase 1]{\includegraphics[width=0.4\columnwidth]{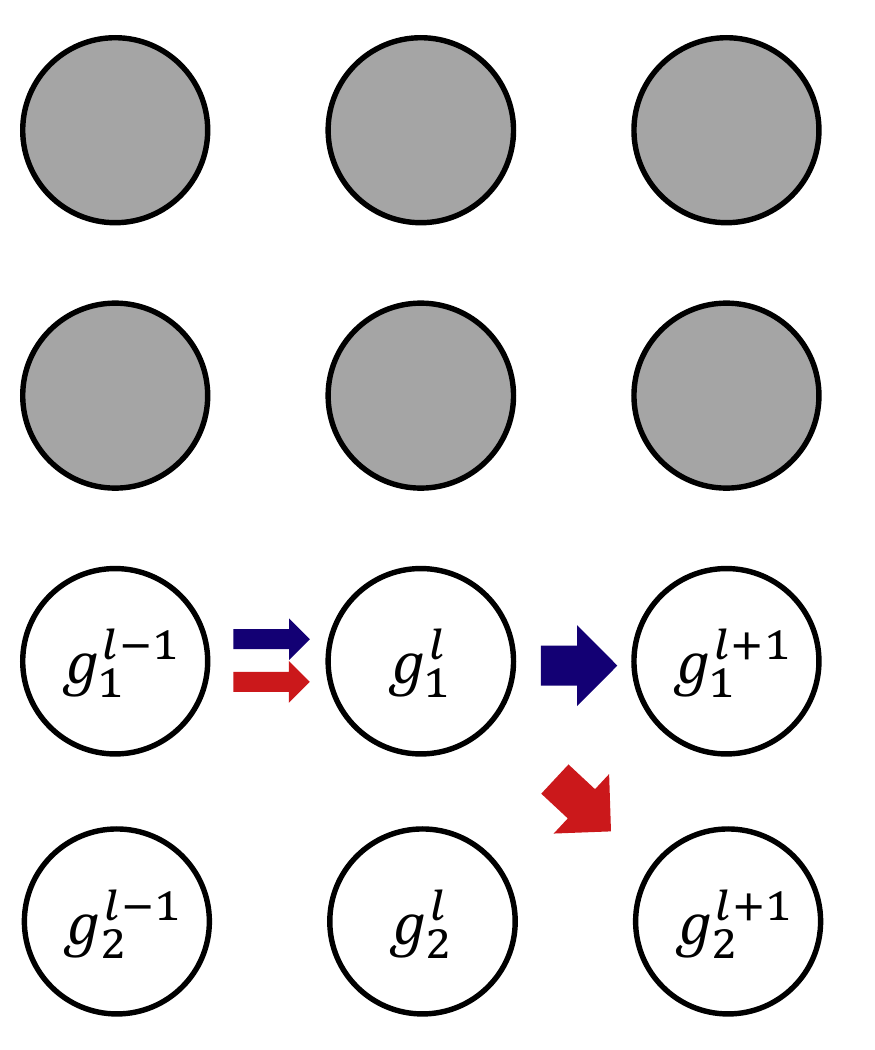}  
	\label{fig:blockdes_2}}}
	\hspace{0.5in}
	{\subfigure[Data-aware progressive training at phase 2]{\includegraphics[width=0.4\columnwidth]{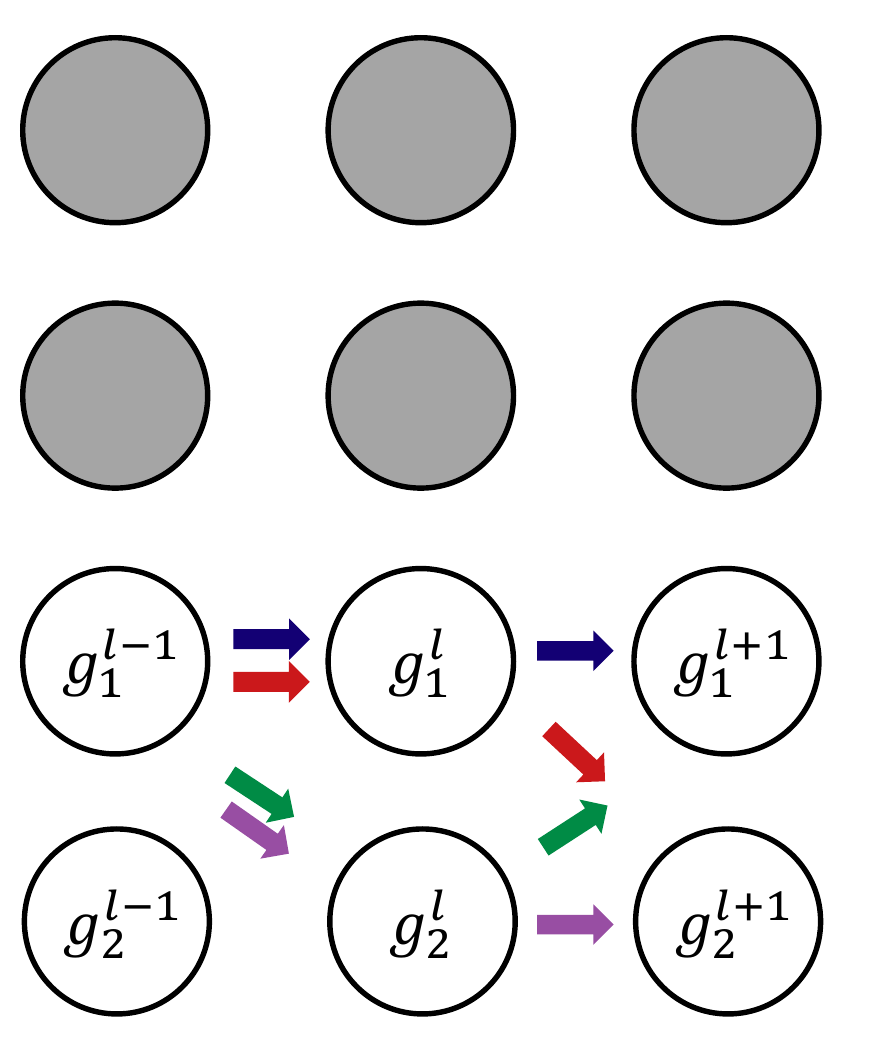}  
	\label{fig:progressive}}}
	\caption{Design of SDR block and data-aware progressive training. 
	(a) Illustration of progressive training at phase 1. 
	Each column represents the design of SDR block, which consists of a shared group (2 grey nodes) and several individual groups (2 white nodes).
	Path is defined as the connections of any individual groups in the consecutive blocks.
	In phase 1, we add sub-nets containing blue and red paths.
	(b) Illustration of progressive training at phase 2.  In phase 2, we enlarge the space with sub-nets containing green and purple paths. }
	\label{fig:SDR_fig}
\end{figure}

\textbf{Splits of sub-nets and SDR block.} 
Here we introduce our design of SDR block that is modified from ResNet-block~\cite{he2016deep} and scalable to a large number of sub-nets. 
Without loss of generality, we denote every column in Fig.~\ref{fig:SDR_fig} as a block since our discussion of block behavior is the same as the layer behavior and all layers in the same block perform identically. We split the channels of each block into two parts: individual groups and shared groups. A path is defined as an \textbf{arbitrary} connection of individual groups between consecutive blocks. There are 3 blocks(columns) and each block contains 2 share groups(grey nodes) and 2 individual groups(white nodes). $[g^{l-1}_1,g^l_1,g^{l+1}_1]$, $[g^{l-1}_1,g^l_1,g^{l+1}_2]$ are two example paths showed as the blue and red paths, where $g_i^l$ denotes the $i$-th individual group of the $l$-th block. The total number of paths can be computed from the number of individual groups and the number of blocks, that is $2^3=8$ in the figure. This design makes the model size grow log-linearly with the number of paths, which is extremely space-saving than training a model for one sub-dataset.
In general, each $D_i$ will be mapped to a path in advance.
When data in $D_i$ comes, it will inference the block with the concatenation of the shared group and the individual group defined in the path, thus $W_i$ is defined as the union of parameters in the corresponding path and all shared groups.

\textbf{Data-aware progressive training.}
It is challenging to train a large number of sub-nets simultaneously due to the instability of the training process.
Naively sampling a sub-net and training it with its corresponding dataset always leads to instability, which finally results in feature collapse in self-supervised learning.
We therefore propose the data-aware progressive training by block to stabilize the optimization process. 
A network space is defined and enlarged progressively after each phase. 
At each phase, we only sample and train the networks inside the space. 
Specifically, the network space only contains the largest network at first. 
We start adding sub-nets whose paths only differ in the last block (i.e. the blue and red paths in Fig.~\ref{fig:blockdes_2}).
In the next phase, we continue to add sub-nets with path green and purple, thus paths of all sub-nets in the space now differ in the last two blocks, and go on. 
With such progressive training, we are able to train the largest network and many sub-nets simultaneously.

\textbf{Task-customized knowledge distillation.}
Besides progressive training by blocks, we further propose a task-customized distillation method called SiamKD to balance the model discrepancy and the possible performance drop resulted from training with fewer data and less time.
Specifically, features provided by sub-nets are also applied to predict the features of the largest network.
The loss function is:
\begin{footnotesize}
\begin{equation}
\mathcal{L}_{SiamKD}(D_i; W_i)  = \mathop{\mathbb{E}}_{\substack{x_1, x_2\sim\tau(x)\\ x \sim D_i}} 
\frac{h(f_{W_i}(x_2))}{\|h(f_{W_i}(x_2))\|_2}
\frac{f_{W_0}(x_1)}{\|f_{W_0}(x_1)\|_2} .
\label{eqn:SiamKD}
\end{equation}
\end{footnotesize}

Note that the stop gradient operation is performed on the SDRnet when calculating $\mathcal{L}_{SiamKD}$, as we distill the SDRnet to each sub-net unilaterally.
Experiments show that SiamKD significantly outperforms the L2 distillation loss.
See the ablation study in Sec.~\ref{sec:ablation} for more details.

\subsection{Deployment}
\label{sec:dep}

When a downstream task comes, one can route among all the sub-nets to find the best pre-trained model for the task.
As for classification task, one practical implementation is to adopt the k-nearest-neighbor (kNN)~\cite{wu2018unsupervised} classifier for fast performance evaluation.  
For detection task, early stopping can be applied to choose the best pre-trained model.
Our experimental results in Sec.~\ref{sec:experiment} verify the effectiveness and efficiency of the above model selection procedures.

%% file: experiment.tex
\section{Experiment}

\label{sec:experiment}

\begin{table*}[t]
\centering
   \scalebox{0.95}
{
  {\small{{\setlength{\tabcolsep}{0.7mm}
    \begin{tabular}{lccccccccccccc}
    \toprule
    & Epochs & Aircraft & Caltech & Cars & C10 & C100 & DTD & Flowers & Food & Pets & SUN & VOC & Avg. \\
    \midrule
    Supervised & 90 & 43.59 & 90.18  & 44.92  & 91.42  & 73.90  & 72.23  & 89.93 &  69.49  & 91.45  & 60.49  & 83.60  & 73.75\\
    \midrule
    InsDis~\cite{wu2018unsupervised} & 200  & 36.87  & 71.12  & 28.98  & 80.28  & 59.97  & 68.46  & 83.44  & 63.39  & 68.78  & 49.47  & 74.37  & 62.29 \\
    MoCo-v1~\cite{he2020momentum} & 200 & 35.55  & 75.33  & 27.99  & 80.16  & 57.71  & 68.83  & 82.10  & 62.10  & 69.84  & 51.02  & 75.93  & 62.41 \\
    PIRL~\cite{misra2020self} & 200  & 37.08 & 74.48 & 28.72 & 82.53 & 61.26 & 68.99 & 83.60  & 64.65 & 71.36 & 53.89 & 76.61 & 63.92 \\
    PCL-v1~\cite{li2020prototypical} & 200 & 21.61  & 76.90  & 12.93  & 81.84  & 55.74  & 62.87  & 64.73 & 48.02  & 75.34  & 45.70  & 78.31  & 56.73 \\
    PCL-v2~\cite{li2020prototypical} & 200 & 37.03 & 86.42 & 30.51 & 91.91 & 73.54 & 70.59 & 85.34 & 64.88 & 82.76 & 56.25 & 81.14 & 69.12 \\
    MoCo-v2~\cite{chen2020improved} & 800 & 41.79 & 87.92 & 39.31 & 92.28 & 74.90  & 73.88 & 90.07 & 68.95 & 83.30  & 60.32 & 82.69 & 72.31 \\
    SimCLR-v1~\cite{chen2020simple} & 1000 & 44.90  & 90.05 & 43.73 & 91.18 & 72.73 & 74.20  & 90.87 & 67.47 & 83.33 & 59.21 & 80.77 & 72.59 \\
    SimCLR-v2~\cite{chen2020big} & 800 & 46.38 & 89.63 & 50.37 & 92.53 & 76.78 & 76.38 & 92.90  & 73.08 & 84.72 & 61.47 & 81.57 & 75.07 \\
    InfoMin~\cite{tian2020makes} & 800 & 38.58 & 87.84 & 41.04 & 91.49 & 73.43 & 74.73 & 87.18 & 69.53 & 86.24 & 61.00    & 83.24 & 72.21 \\
    SeLa-v2~(Asano et.al. 2019) & 400  & 37.29  & 87.20  & 36.86  & 92.73  & 74.81  & 74.15  & 90.22  & 71.08  & 83.22  & 62.71  & 82.73  & 72.09 \\
    DeepCluster-v2$ ^*$ \cite{caron2018deep} & 400 & 48.75 & \underline{90.52} & 50.94 & \underline{94.15} & \underline{79.33} & \underline{76.70} & 93.98 & 75.90 & 86.78 & \textbf{65.41} & 84.30 & 76.98 \\
    SwAV$^*$~\cite{caron2020unsupervised} & 400 & 51.37 & 89.65 & 52.59 & 93.39 & 78.72 & \textbf{78.09} & 93.94 & \underline{75.92} & 86.81 & 63.55 & 83.92 & 77.09 \\
    \midrule
    SimSiam$^{**}$~\cite{chen2020exploring} & 200 & 51.30 & 87.02  & 53.80 & 89.12  & 68.43 & 72.99 & 91.83 & 67.35 & 83.64 & 52.97 & 83.40  & 72.90 \\

    SDR (SimSiam) & 200 & \textbf{55.84} & 87.55 & \textbf{61.06} & 90.27 & 71.39 & 74.47 & 92.61 & 68.93 & 85.03 & 55.89 & \underline{85.02} & 75.28$^{+2.38}$ \\
    \midrule
    BYOL$^{**}$~\citep{byol20} & 200 & 45.46 & 87.82 & 45.91 & 91.42 & 74.37 & 73.14 & 90.95 & 73.13 & 84.62 & 56.43 & 81.99 & 73.20 \\
    
    BYOL$^{**}$~\citep{byol20} & 400 & 48.93 & 90.39 & 54.43 & 92.12 & 75.97 & 76.65 & \underline{94.50}  & 74.13 & \underline{87.81} & 57.99 & 82.48  & 75.95 \\
    
    SDR (BYOL) & 400 & \underline{52.51} & \textbf{91.12} & \underline{56.09} & \textbf{94.27} & \textbf{79.90} & 76.33& \textbf{94.75} & \textbf{76.98} & \textbf{89.86} & \underline{63.62} & \textbf{85.12} & \textbf{78.23}$^{+2.28}$  \\
    \bottomrule
    
    \end{tabular}}}}}
    \caption{Transfer performance(\%) of self-supervised pre-training models on various classification downstream tasks (Bold: best, underline: second best). Supervised baseline is also provided in the first row. SDR improves the baselines significantly by 2.38\% and 2.28\%. Especially, SDR(BYOL) performs best on seven tasks and second-best on three tasks, achieving state-of-the-art averaged accuracy.
$^*$: we take the officially released pre-trained weights and report the transfer performance.
$^{**}$: denotes our re-implementation under the same training epochs with SDR for a fair comparison.} 
  \label{tab:linearres}
\end{table*}

In this section, we apply the proposed SDR to train SDRnet and a series of sub-nets.
We demonstrate the effectiveness of SDR by evaluating the resulting pre-trained models on various downstream tasks including classification and detection. We also take ablation studies on the number of sub-nets, training time and the distillation method as shown in Sec.~\ref{sec:ablation}.

\subsection{Implementation Details}
\label{sec:exp details}

\textbf{Model configuration.} We apply the SDR block in all four stages of ResNet. In each stage, all blocks have four individual groups and one shared group. The size of shared groups is half of all groups. 
All blocks in same stage perform identically. So we can generate $4^4=256$ different sub-nets. 
For comparison, we enlarge our model so that the size of each sub-net is close to that of ResNet-50~\citep{he2016deep}, the most commonly used backbone in SSL.
For deployment, we reset each sub-net with the corresponding batch normalization (BN) statistics in pre-training following~\cite{cai2019once}. 
We adopt ImageNet as the dataset for self-supervised pre-training without using labels. We use SimSiam~\citep{chen2020exploring} and BYOL~\citep{byol20} as our baseline models. Considering the simplicity and effectiveness, we perform most ablations on SimSiam.

\textbf{Downstream tasks.} 
We validate our method on both classification and detection.
For \textbf{classification tasks}, we adopt the benchmark proposed in \cite{ericsson2020well}, which considers 11 datasets including both coarse-grained (\emph{e.g.}, CIFAR100 and VOC2007) and fine-grained ones (\emph{e.g.}, Standard Cars and FGVC Aircraft), as detailed in Appendix A.
The quality of the pre-trained representations is evaluated by training a supervised linear classifier upon the \textbf{frozen} representations in the training set, and then testing it in the validation set. 
For \textbf{detection task}, we evaluate the pre-trained models on PASCAL VOC detection dataset with Faster-RCNN, following the transfer protocol of MoCo~\cite{chen2020improved}.
Specifically, the pre-trained model is fine-tuned on the VOC \texttt{trainval07+12} set and evaluated on the VOC \texttt{test2007} set.
See Appendix A for more experimental details and hyper-parameters.

\subsection{Results and Analysis}

\textbf{Classification.} The transfer performance of pre-trained models on classification tasks are summarized in Table~\ref{tab:linearres}.
As can be seen, SDR improves the performance on all the downstream datasets, compared with the model pre-trained on full ImageNet, i.e., the SimSiam and BYOL baselines.
SDR achieves 2.38\% and 2.23\% improvement of accuracy respectively over eleven downstream tasks, demonstrating the effectiveness of task-customized pre-training to alleviate negative transfer. 
Especially, SDR(BYOL) reaches the best performance on 7 tasks and second best on 3 tasks, whose average accuracy also outperforms other state-of-the-art methods.

Note that the baseline SimSiam and BYOL uses ResNet-50 as the backbone, whose parameter count is 23.5 million, while the size of each SDR sub-net is 22.6 million. With a smaller model, we achieve better results. Besides, under the same time consumption, we are able to train 256 sub-nets, showing the scalability of our method. In terms of efficient deployment, it takes few minutes to route among all sub-nets using the kNN classifier to find the best model.
Compared with the total training time of SDR, which usually takes hundreds of hours, the searching time is negligible.
On Food-101~\citep{bossard14}, for example, it takes 20 minutes with 8*V100 to decide the best route.

\begin{figure}[tbp] 
	\centering
	{
	\subfigure[Performance gains of sub-nets ]{\includegraphics[width=0.45\columnwidth ]{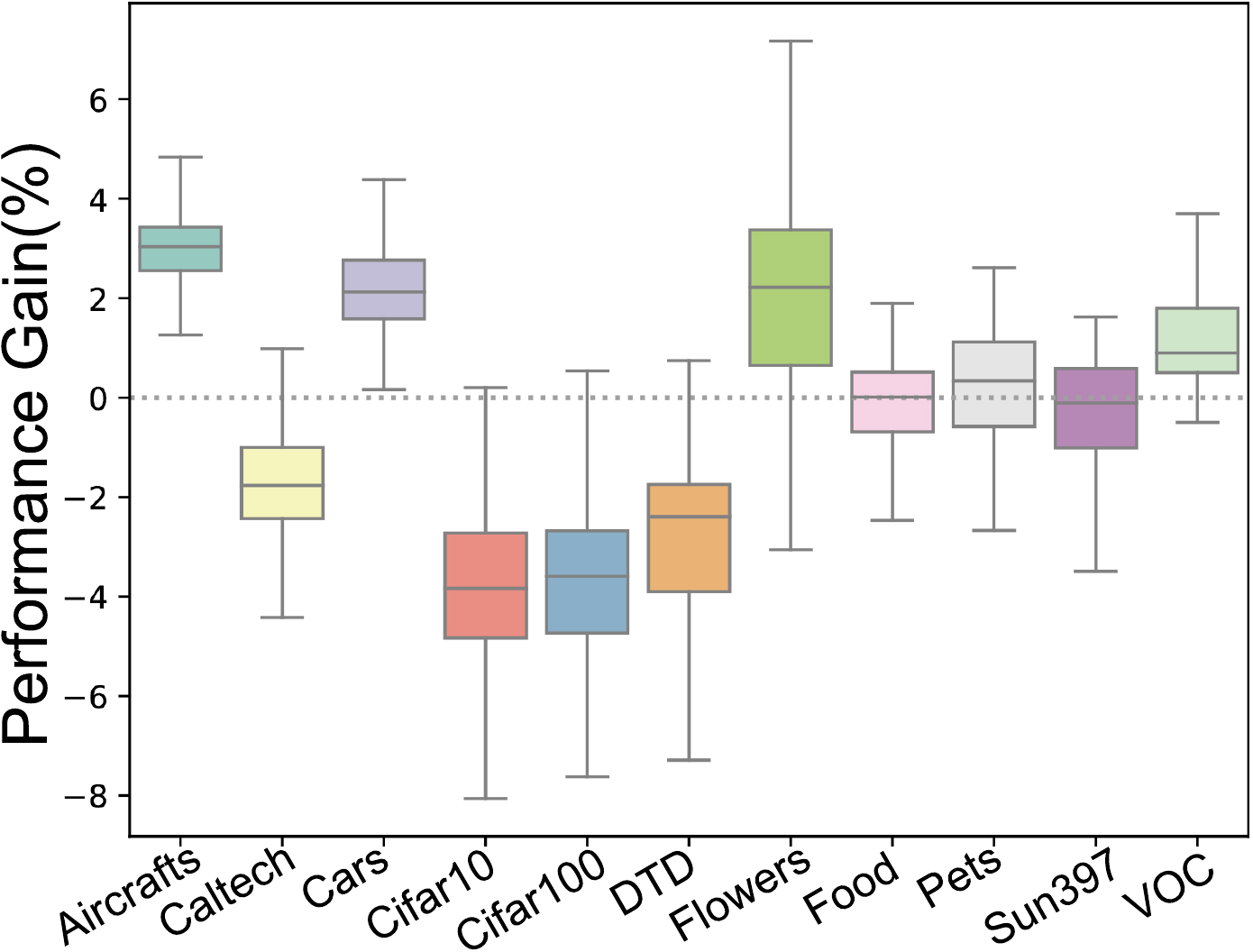}  \label{fig:2a}}}
	\hspace{0.01in}
	{
	\subfigure[Histogram of performance gains on Aircraft]{\includegraphics[width=0.46\columnwidth,]{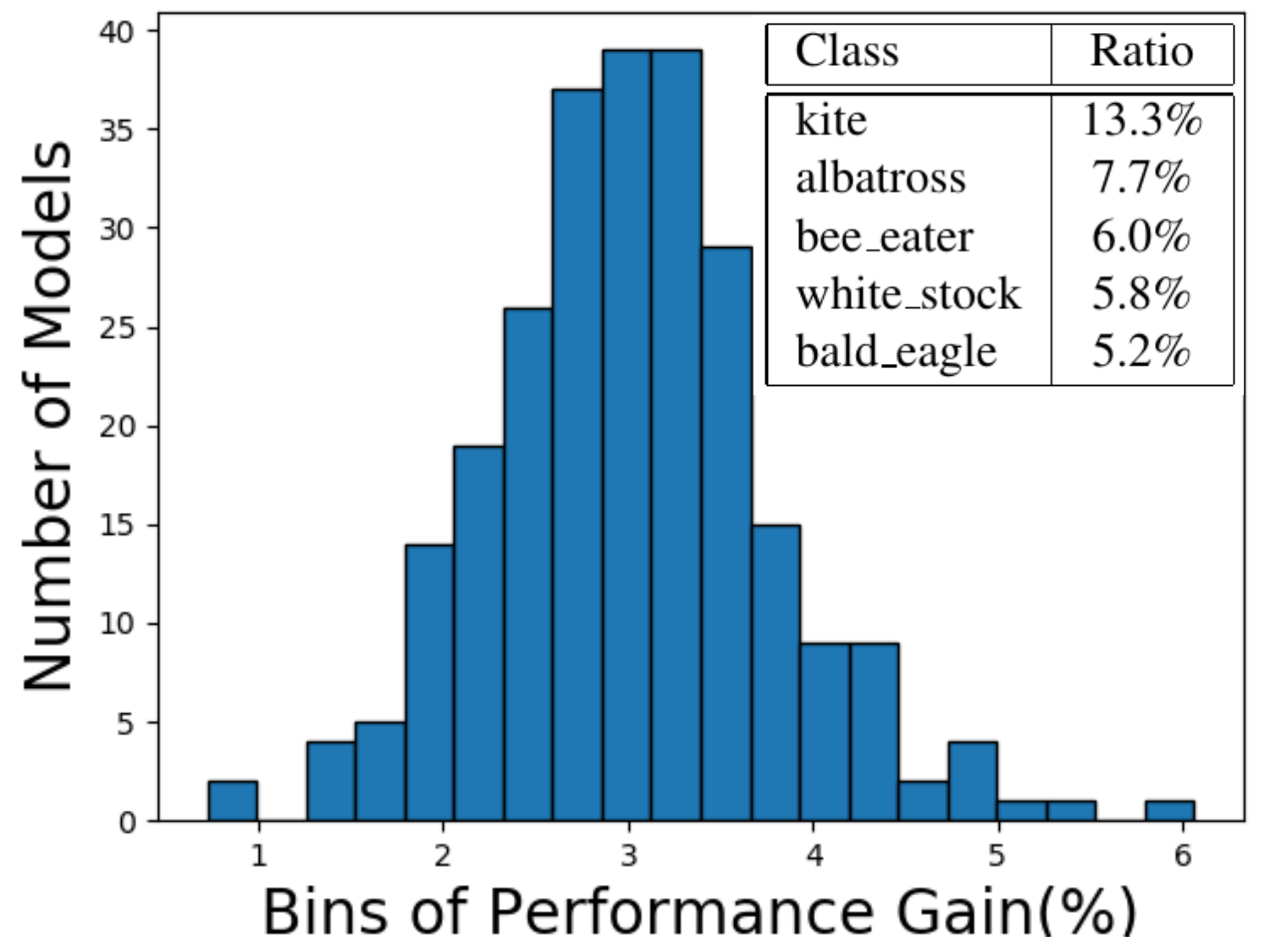}  
	\label{fig:2b}}}
	\caption{(a) Performance gain on kNN accuracy of the 256 sub-nets pre-trained by SDR compared with the baseline trained on the full ImageNet. (b) Histogram of the performance gains on Aircraft dataset. The $x$-axis is the performance gain on kNN accuracy compared with the baseline, and the $y$-axis is the number of models.
	}
\end{figure}

\textbf{Analysis on downstream tasks.} 
We notice that the performance of the sub-nets varies significantly for different downstream datasets. As showed in Fig.~\ref{fig:2a}, we provide the performance gains on kNN accuracy of the 256 sub-nets compared with baseline trained on full ImageNet.
The downstream tasks including Aircraft, Cars and Flowers have significant average performance improvement when using SDR.
That might be because these datasets are fine-grained datasets sensitive to the negative transfer.
Therefore, a subset of ImageNet tends to provide a better pre-trained model.
On the other hand, downstream tasks like CIFAR10, CIFAR100 and DTD show limited improvement when using SDR.
That might be because these datasets contain classes similar to those in ImageNet,
so that effects of negative transfer are negligible.
As a result, the full ImageNet may provide more applicable pre-trained models. 
These observations are also consistent with the preliminary experiments (see Sec.~\ref{sec:pre}).

For illustration purpose, we plot the histogram of kNN accuracy on the Aircraft dataset over the 256 sub-nets.
The results are summarized in Fig.~\ref{fig:2b}.
We further investigate the distribution of classes for the subset that results in the best kNN accuracy on Aircraft.
As can be seen, the best pre-trained model for Aircraft is actually pre-trained on a subset of ImageNet mainly containing images of flying objects, such as kite, albatross and stork.
The results indicate the effectiveness of the data clustering and data-aware progressive training process.

\begin{table}[t]
  \centering

  \centering
     \small{\scalebox{0.95}{\setlength{\tabcolsep}{1.5mm}{
      \begin{tabular}{lccc}
    \toprule
    & AP & AP50 & AP75\\
    \midrule
    Supervised  & 53.26  & 81.51  & 59.07\\
    \midrule
    InsDis   & 48.82  & 76.43  & 52.40\\
    MoCo-v1   & 50.51  & 78.06  & 54.55\\
    PIRL   & 45.08  & 72.50  & 47.80\\
    PCL-v1   & 53.93  & 81.69  & 59.33\\
    PCL-v  & 53.92  & 81.89  & 59.35\\
    MoCo-v2   & 44.74  & 72.82  & 47.01\\
    SimCLR-v1   & 52.19  & 81.36  & 56.92\\
    SimCLR-v2   & 51.42  & 79.40  & 55.89\\
    InfoMin   & 44.92  & 72.72  & 47.41\\
    SeLa-v2   & 50.41  & 80.55  & 54.35\\
    DeepCluster-v2 & 51.03 & 80.93 & 55.51 \\
    SwAV & 52.07 & 81.50 & 56.03 \\
    \midrule
    SimSiam & 54.17 & 80.09 & 59.58 \\
    SDR (SimSiam) & \textbf{55.65$^{+1.48}$} & 81.16$^{+1.07}$  &  \textbf{60.89$^{+1.31}$} \\
    \midrule
    BYOL & 52.75 & 81.83 & 58.35 \\
    SDR (BYOL) & 53.53$^{+0.78}$ & \textbf{82.69$^{+0.86}$} & 59.25$^{+0.90}$ \\
    \bottomrule
    \end{tabular}}}}
      \caption{Detection transfer results(\%) from pre-trained models using Faster R-CNN FPN on PASCAL VOC. The models are trained with all layers fine-tuned. Metrics including AP, AP50 and AP75 are reported.}
  \label{tab:detection}%
\end{table}%

\begin{table*}[t]
  \centering
 {\small{
 \scalebox{0.95}{ \setlength{\tabcolsep}{1.5mm}{
    \begin{tabular}{lcccccccccccccccc}
    \toprule
      & Dataset & Param \# & Aircraft & Caltech & Cars & C10 & C100 & DTD & Flowers & Food & Pets & SUN & VOC & Avg. \\
    \midrule
    SimSiam & IN & 23.5M & 51.30  & 87.02 & 53.80  & 89.12 & 68.43 & 72.99 & 91.83 & 67.35 & 83.64 & 52.97 & 83.40  & 72.90 \\
    SDR & IN & 22.6M & 51.95 & 86.79 & 55.62 & 88.60  & 67.54 & 72.09 & 91.83 & 67.66 & 82.44 & 51.58 & 81.42 & 72.50 \\
    SDR & Random & 22.6M & 48.21 & 86.23 & 50.25 & 86.62 & 64.41 & 72.29 & 92.47 & 64.45 & 82.21 & 50.39 & 80.13 & 70.70 \\
    \midrule
    SimSiam & IN & 56.3M & 52.31 & 87.46 & 54.50  & 90.05 & 68.75 & 73.19 & 93.11 & 68.56 & 83.34 & 53.27 & 83.90  & 73.49 \\
    SimSiam & IN & 22.6M & 43.82 & 63.00 & 37.72 & 80.51 & 50.10  & 62.81 & 73.64 & 50.24 & 61.43 & 32.72 & 68.62 & 56.78 \\
    \midrule
    SDR & Cluster & 22.6M & 55.84 & 87.55 & 61.06 & 90.27 & 71.39 & 74.47 & 92.61 & 68.93 & 85.03 & 55.89 & 85.02 & \textbf{75.28} \\
    \bottomrule
    \end{tabular}}}}%
    \caption{Effects of clustering and progressive training. The second column(Dataset) for all SDR models means the dataset used for training each sub-net. SimSiam is always trained by total ImageNet. The third column(Param \#) means the parameter count of the model during testing time. The last line SDR-Cluster-22.6M is our proposed model.
  (1) The first section is the comparison of the models trained by the total ImageNet, random splits and the clusters.
  (2) Comparison of the lottery ticket theorem~\cite{frankle2018lottery} is provided in section 2. SimSiam is pre-trained under 56.3M and then pruned to 22.6M, which are the exact sizes of the super-net and sub-net in our SDR framework.
}
  \label{tab:clustering}%
  }
\end{table*}%

\begin{table}[ht]
  \centering

  \small{\setlength{\tabcolsep}{0.8mm}{
    \begin{tabular}{cccccc}
    \toprule
        \# of sub-nets      & 1     & 4    & 16  & 64   & 256 \\
    \midrule
        kNN accuracy & 53.42 & 55.77 & 56.83 & 57.83 & 58.13 \\
        Training time (GPU hours) & 260   & 370 & 400 &  460  & 500 \\
    \bottomrule
    \end{tabular}}}
      \caption{Results of kNN accuracy and training time (i.e., GPU hours) for SDR with different number of sub-nets.}
  \label{tab:num_subnets}
\end{table}

\textbf{Detection.} The transfer results for detection task are provided in Table~\ref{tab:detection}. For detection task, SDR also improves the baselines by 1.48\% and 0.78\% in AP with smaller model size, compared with the models pre-trained on the full ImageNet, which further verifies the necessity of using task-customized pre-trained models. 
In detection, we adopt fast deployment through early stopping. We train the model that performs best at iteration 1000, which takes about 15 minutes on 8*V100 for each model. Compared with the six-hours' fine-tuning with 8*V100, the routing procedure takes much less time to produce a reasonable model.

\subsection{Ablation Study}
\label{sec:ablation}

\textbf{Effects of clustering.}
Here we analyze the importance of clustering through two controlled experiments. We first train each sub-net with the \textbf{total ImageNet(IN)}, with all other modules unchanged, including progressive training and knowledge distillation. We call this model SDR-IN-22.6M(line 2 of Table~\ref{tab:clustering}) and name our original model as SDR-Cluster-22.6M(last line of the table). 
SDR-Cluster-22.6M outperforms SDR-IN-22.6M consistently among all tasks. We achieve a nearly 2.7\% mean accuracy improvement, which indicates that training with separate subsets contributes a lot to our SDR. Note that without using clustered subsets, SDR-IN-22.6M is even exceeded by the SimSiam baseline, suggesting that the usage of clustered subsets is the most crucial component in the SDR framework. 
The other experiment is to train the model with random clusters. Specifically, we split the ImageNet \textbf{randomly} into 256 sub-datasets and perform our training procedure subsequently, named as SDR-Random-22.6M(line 3), which shows a degenerated performance over all tasks, indicating the great importance of a reasonable clustering method.

\begin{figure}
  \centering
	\includegraphics[width=0.5\columnwidth]{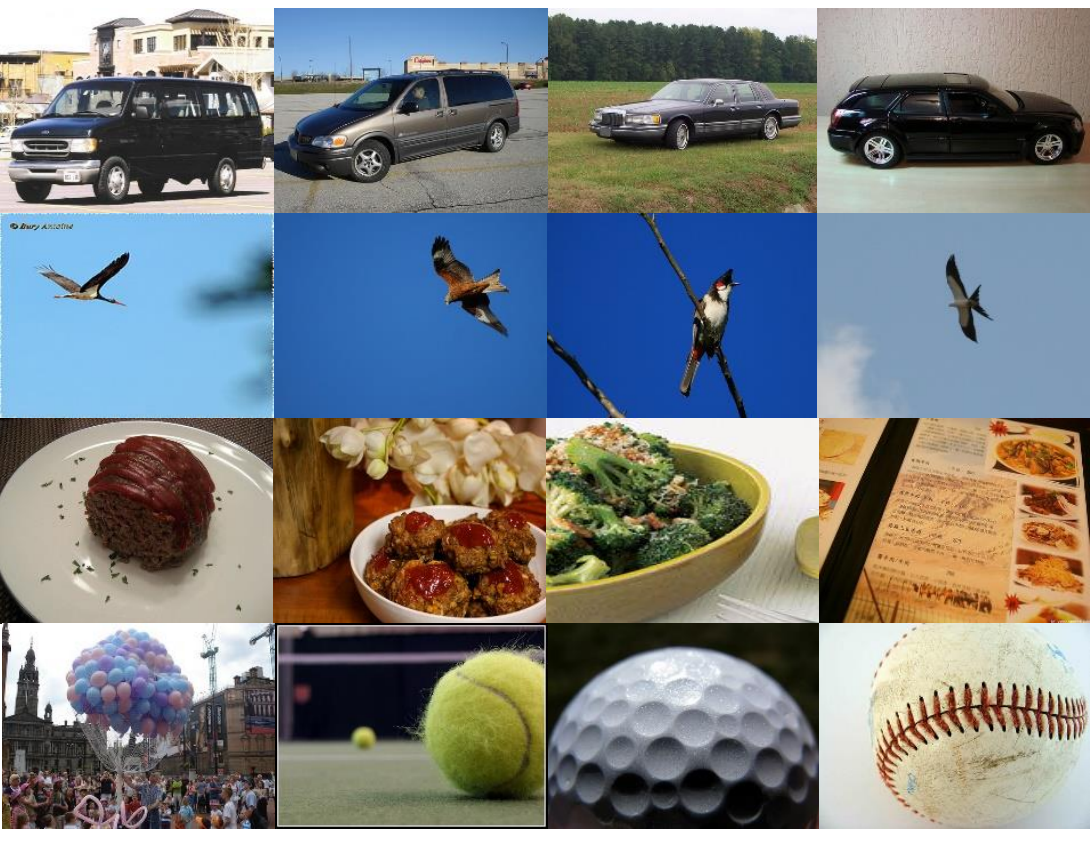}
	\caption{Image samples of different data clusters.}
  \label{fig:cluster}
\end{figure}

\textbf{Visualization of clustering.} To visualize the clustering procedure, we single out the four clusters that provide the best transfer performance on Standard Cars, FGVC Aircraft, Food-101 and DTD, respectively.
We randomly plot four samples of each cluster, as shown in each row of Fig.~\ref{fig:cluster}.
As can be seen, images in a cluster have similar semantics, suggesting the effectiveness of data clustering.
These images also have semantics similar to the corresponding downstream tasks, which brings improvement of the transfer performance.

\textbf{Effects of progressive training v.s. lottery ticket theorem}~\cite{frankle2018lottery}, which suggests that neural networks might rely on internal sub-nets that are significantly smaller than the full parameter count for the majority of their predictive accuracy. 
In this experiment, we try to make lottery ticket theorem explicit to see how exactly the usage of sub-nets may contribute to the success of SDR.
We first train a SimSiam with the same amount of parameters with our SDR super-net whose size is 56.3M, denoted as SimSiam-IN-56.3M. 
Then we perform pruning, the method to get effective sub-nets used in \cite{frankle2018lottery}, to get SimSiam-IN-22.6M, the `winning ticket' of SimSiam-IN-56.3M. 
Here 22.6M is the exact size of SDR sub-net used for each downstream task.
As shown in Table~\ref{tab:clustering}, SDR-Cluster-22.6M outperforms SimSiam-IN-22.6M dramatically and consistently among all tasks, which indicates that choosing the proper sub-dataset is more crucial than getting the `winning ticket' of the large model. Furthermore, we notice a large performance drop of SimSiam after pruning, while SDR performs even better than the large model SimSiam-IN-56.3M, demonstrating that simply pruning is not enough to get a better sub-net while SDR is more effective to get better performance. Based on the experiments above, we tend to believe that SDR indeed benefits from sub-datasets other than merely making the lottery ticket theorem explicit.

\textbf{Number of sub-nets.}
We analyze how different numbers of sub-nets affect the final results by evaluating the kNN accuracy averaged over 11 downstream tasks. 
The model with one sub-net is actually the SimSiam baseline.
The kNN accuracy and the corresponding training time are reported in Table~\ref{tab:num_subnets}.
With a larger number of sub-nets, the kNN accuracy increases significantly. 
Intuitively, a larger number of sub-nets tends to have a higher probability of providing proper sub-sets for various downstream tasks, yet inevitably requiring a longer training time.
The proposed SDR, however, only introduces moderate extra training time with the increasing number of sub-nets, which is applicable for real applications. 

\textbf{Effects of distillation.} We compare our task-customized knowledge distillation method, SiamKD, with the vanilla L2 distillation loss~\citep{hinton2015distilling}.
We train SDRnet with the L2 distillation loss and SiamKD in Eqn.~(\ref{eqn:SiamKD}), respectively, following the implementation in Sec.~\ref{sec:exp details}.
For the 11 downstream classification tasks, we compute the average kNN accuracy of the best sub-net, as well as the average standard deviation of the 256 sub-nets for each downstream task.
The accuracy of SiamKD is $58.13\pm2.38$, while L2 loss only gets $54.66\pm0.07$.
In addition to the inferior performance, the L2 distillation results in a small standard deviation of kNN accuracy and homogenized sub-nets, while SiamKD maintains the feature diversity of sub-nets, which is essential for providing a task-customized model. 
In practice, we also find SiamKD helps to provide better feature representations and stabilize the training process.

%% file: conclusion.tex
\section{Conclusion}

In this work, we first identify the negative transfer phenomenon in SSL that involving semantic-irrelevant data in pre-training may degenerate the downstream performance. To address this issue, we propose a novel task-customized SSL paradigm called Scalable Dynamic Routing (SDR). 
SDR first cluster the training data and then train each sub-net with a different cluster through a data-aware progressive training framework. Finally, customized sub-nets are deployed to different downstream tasks efficiently.
In the experiments, we succeed in training 256 sub-nets simultaneously, with a total training cost less than twice of the SSL baseline that provides only one pre-trained model, achieving SOTA results on average accuracy among 11 downstream classification tasks and AP on PASCAL VOC detection task.

%% file: appendixA.tex
\section{Experimental Details}
\label{app:exp}

\textbf{Classification Task.}
The models are tested under 11 downstream tasks, which consist of Food-101~\citep{bossard14}, CIFAR10~\citep{krizhevsky2009learning}, CIFAR100~\citep{krizhevsky2009learning},  SUN397~\citep{xiao2010sun}, Standard Cars~\citep{krausecollecting}, FGVC Aircraft~\citep{maji2013fine}, VOC2007~\citep{everingham2010pascal}, DTD~\citep{cimpoi2014describing}, Oxford-IIIT Pets~\citep{parkhi2012cats}, Caltech-101~\citep{fei2004learning} and Oxford 102 Flowers~\citep{nilsback2008automated},
following~\cite{ericsson2020well}, including the train-test splits and evaluation metrics.
Two criteria are adopted to evaluate our model: kNN accuracy and many-shot classification protocol (also known as linear protocol).
When the model is evaluated in kNN, the images are resized to 256 along the shorter side followed by a center crop of 224x224. 
Each label of the test images is estimated from the $k$ nearest neighbors of training samples in the feature space, 
where $k$ is set to be 200 by default. 
kNN criterion is used in routing for efficient deployment and our ablation of the number of sub-nets.
For the many-shot classification protocol, same augmentation as kNN criterion is performed. The linear layer is constructed behind the frozen backbone. Following~\cite{ericsson2020well}, we select the weight of $l_2$ regularisation on the validation set and optimise the model with L-BFGS on the softmax cross-entropy objective. We follow their public code to evaluate our models.

\textbf{Detection Task.} 
The train set of our detection task is VOC \texttt{trainval07+12} and the test set is VOC \texttt{test2007}. 
The entire network is optimized following our finetune setting.
The features are extracted from the backbone followed by a Faster R-CNN detector head for prediction. Synchronized BN is applied among all models. During training, the images are resized that the shorter side is one of [80, 512, 544, 576, 608, 640, 672, 704, 736, 768, 800] and 800 pixels for testing.
The model is trained for 24k iteration with warm-up of 100 iterations. 
The initial learning rate is 0.02 which is decayed by a factor of 10 at the iteration of 18k and 22k. 
Other training details follow the default settings of detectron2~\cite{wu2019detectron2}.

%% file: appendixB.tex
\section{Splits of ImageNet in Preliminary}
\label{app:split of IN}
In the preliminary experiment, we split the ImageNet into two sub-datasets based on semantic dissimilarity in WordNet Tree. 
Specifically, we search each class in WordNet and let those classes that share the same ancestor at depth four be in the same sub-dataset.
Here we illustrate some classes of each sub-datasets.

\textbf{Subset-A.} The number of classes in Subset-A is 520. The classes are: abacus;
abaya;
academic gown, academic robe, judge's robe;
accordion, piano accordion, squeeze box;
acoustic guitar;
aircraft carrier, carrier, flattop, attack aircraft carrier;
airliner;
airship, dirigible;
altar;
ambulance;
amphibian, amphibious vehicle;
analog clock;
apiary, bee house;
apron;
ashcan, trash can, garbage can, wastebin, ash bin, ash-bin, ashbin, dustbin, trash barrel, trash bin;
assault rifle, assault gun;
backpack, back pack, knapsack, packsack, rucksack, haversack;
bakery, bakeshop, bakehouse;
balance beam, beam;
balloon;
and so on. Most of them are inanimate objects.

\textbf{Subset-B.} There are totally 480 classes in Subset-B, e.g. tench, Tinca tinca;
goldfish, Carassius auratus;
great white shark, white shark, man-eater, man-eating shark, Carcharodon carcharias;
tiger shark, Galeocerdo cuvieri;
hammerhead, hammerhead shark;
electric ray, crampfish, numbfish, torpedo;
stingray;
cock;
hen;
ostrich, Struthio camelus;
brambling, Fringilla montifringilla;
goldfinch, Carduelis carduelis;
house finch, linnet, Carpodacus mexicanus;
bulbul;
jay and so on. Most of them are living things.

%% file: appendixC.tex
\section{More Examples of Clusters}
\begin{figure*}[htbp]
    \centering
    \subfigure[Left: Histogram of performance gains on FGVC Aircraft. Right: More image samples of the best cluster.]{\includegraphics[height=3.1cm]{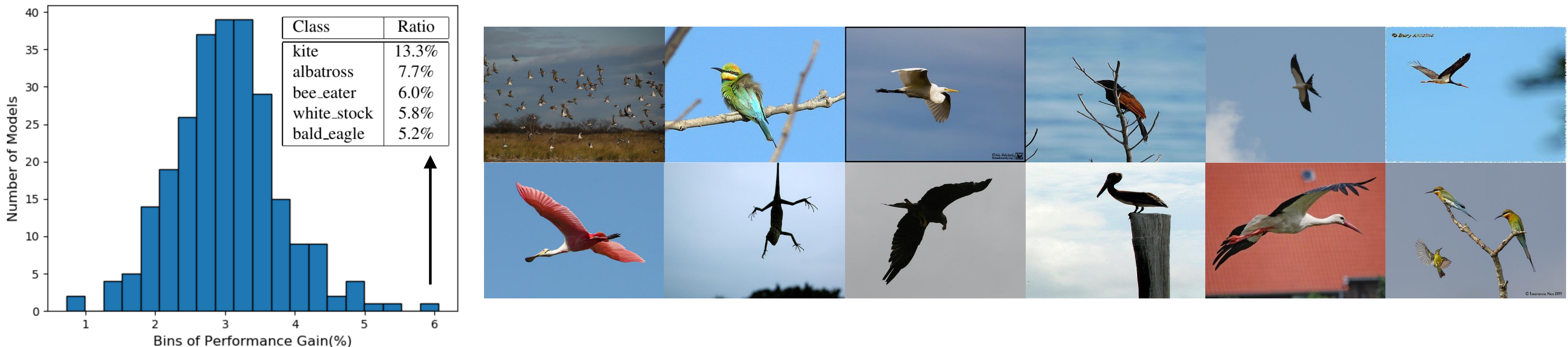}\label{fig:sample_a}}
    
    \subfigure[Left: Histogram of performance gains on Standard Cars. Right: More image samples of the best cluster.]{\includegraphics[height=3.1cm]{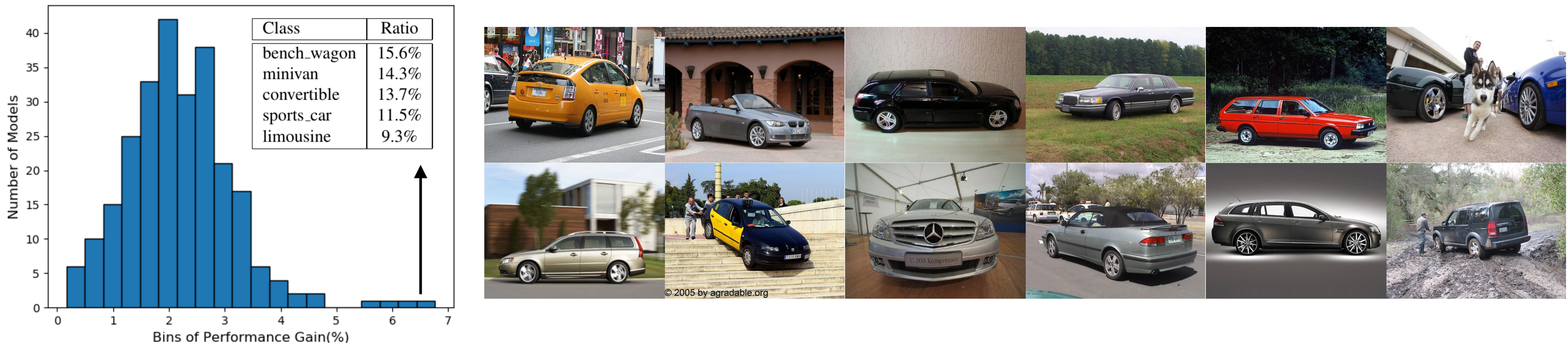}\label{fig:sample_b}}
    
    \subfigure[Left: Histogram of performance gains on Food-101. Right: More image samples of the best cluster.]{\includegraphics[height=3.1cm]{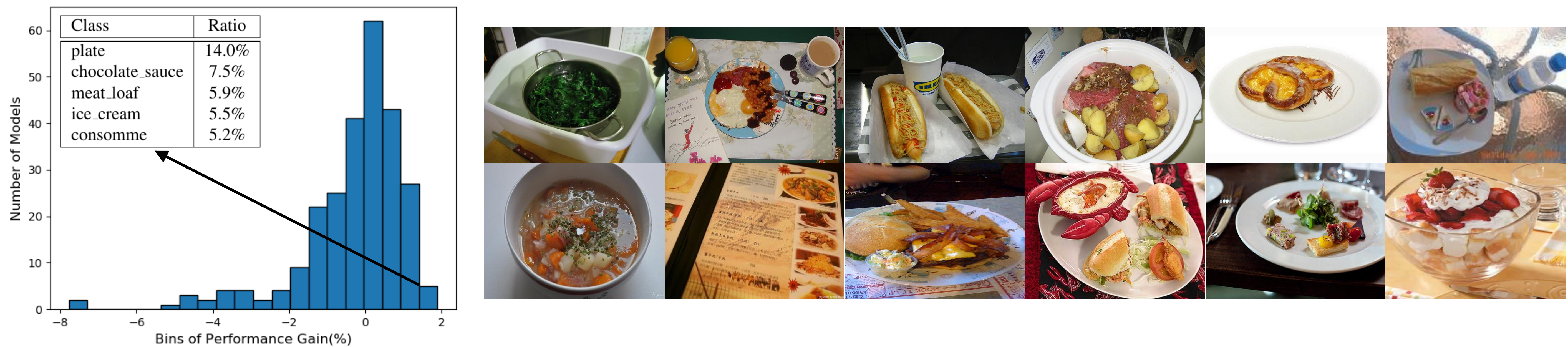}\label{fig:sample_c}}
    
    \subfigure[Left: Histogram of performance gains on DTD. Right: More image samples of the best cluster.]{\includegraphics[height=3.1cm]{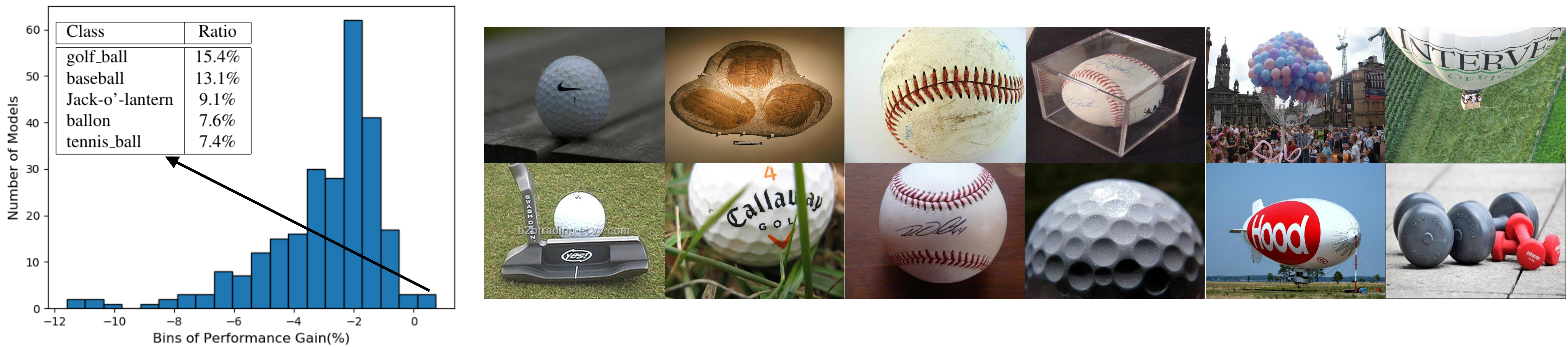}\label{fig:sample_d}}
    
    \subfigure[Left: Histogram of performance gains on Caltech-101. Right: More image samples of the best cluster.]{\includegraphics[height=3.1cm]{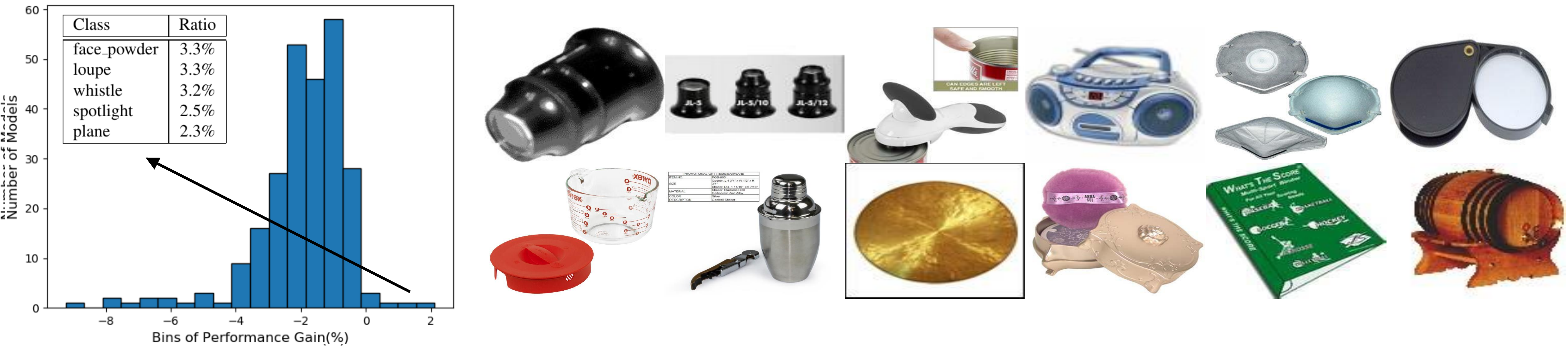}\label{fig:sample_e}}
    
    \subfigure[Left: Histogram of performance gains on CIFAR10. Right: More image samples of the best cluster.]{\includegraphics[height=3.1cm]{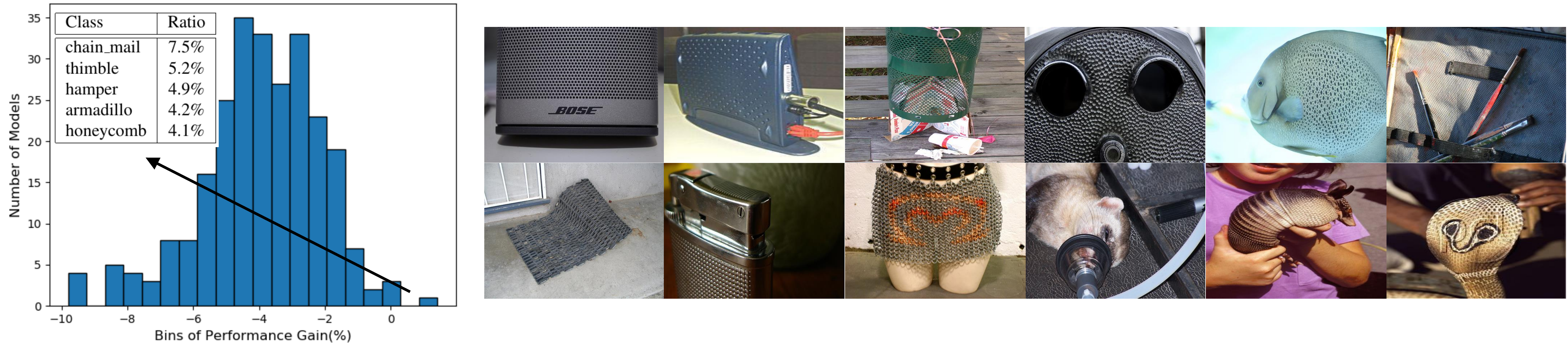}\label{fig:sample_f}}

    \label{fig:splits}
\end{figure*}

Here we illustrate the histogram of performance gains on different downstream tasks, along with the ratio of top-5 most frequent classes in the cluster that transfer best to the corresponding task. We also randomly sample 12 images in each cluster to show the empirical distribution. As shown in the figure below, each cluster contains images with similar semantics or appearance. 
Moreover, the downstream tasks like FGVC Aircraft, Standard Cars and Food-101, which are exactly fine-grained datasets, always perform the best with the closest cluster. 
Differently, datasets like Caltech-101 and CIFAR10 that cover a wide range of semantics need clusters that distribute more uniformly among different classes (see the ratio of each class in Fig.~\ref{fig:sample_e} and~\ref{fig:sample_f}), which follows the conclusion of our preliminary experiment. 
Another finding is that few clusters transfer well to DTD, a dataset for classifying the texture. 
A possible reason is that no semantic-similar clusters can be found in the pre-train datasets, which inspires our future work on more dimensions to cluster the pre-train datasets        , e.g. texture.

%% file: appendixD.tex
\section{Training Procedure of SDR.}
\begin{algorithm}[h]
\caption{Training Procedure of SDR.}
\label{alg:A}
\begin{algorithmic}[1]
\REQUIRE  $k$: the number of sub-nets, $D_{0}$: the entire training dataset
\STATE Split $D_{0}$ into $k$ subsets $D_1, D_2, ..., D_k$.
\STATE Initialize SDRnet with random parameters.
\STATE Initialize network space $N=\left\{ W_0\right\}$.
\REPEAT 
\REPEAT
\STATE Randomly sample a network $W_i$ in $N$.
\STATE Find the corresponding sub-dataset $D_i$ by $W_i$.
\STATE Randomly sample a batch of data in $D_i$.
\STATE Calculate $\mathcal{L}_{SSL}(D_i ; W_i) + \mathcal{L}_{SiamKD}(D_i; W_i)$.
\STATE Optimize $W_i$.
\UNTIL{Converge.}
\STATE Update $N$ with more sub-nets according to data-aware progressive training.
\UNTIL{All sub-nets are trained to converge.}
\end{algorithmic}
\end{algorithm}